\pdfoutput=1

\documentclass[11pt]{article}

\usepackage[]{ACL2023}

\usepackage{times}
\usepackage{latexsym}

\usepackage[T1]{fontenc}

\usepackage[utf8]{inputenc}

\usepackage{microtype}

\usepackage{inconsolata}

\usepackage{graphicx}
\usepackage{subfigure}
\usepackage{listings}
\usepackage{xcolor}
\usepackage{array}
\usepackage{hyperref}
\usepackage{comment}
\usepackage{amsmath}
\usepackage{tikz}
\usepackage{booktabs}

\newcommand{\boldcolor}[2]{{\bfseries \color{#1} #2}}

\definecolor{codegreen}{rgb}{0,0.6,0}
\definecolor{codegray}{rgb}{0.5,0.5,0.5}
\definecolor{codepurple}{rgb}{0.58,0,0.82}
\definecolor{backcolour}{rgb}{0.95,0.95,0.92}

\definecolor{carminered}{rgb}{1.0, 0.0, 0.22}
\definecolor{applegreen}{rgb}{0.55, 0.71, 0.0}
\definecolor{cadmiumorange}{rgb}{0.93, 0.53, 0.18}
\definecolor{babyblueeyes}{rgb}{0.63, 0.79, 0.95}
\definecolor{brightlavender}{rgb}{0.75, 0.58, 0.89}

\lstdefinestyle{mystyle}{
    backgroundcolor=\color{backcolour},   
    commentstyle=\color{codegreen},
    keywordstyle=\color{magenta},
    numberstyle=\tiny\color{codegray},
    stringstyle=\color{codepurple},
    basicstyle=\ttfamily\footnotesize,
    breakatwhitespace=false,         
    breaklines=true,                 
    captionpos=b,                    
    keepspaces=true,                 
    numbers=left,                    
    numbersep=5pt,                  
    showspaces=false,                
    showstringspaces=false,
    showtabs=false,                  
    tabsize=2,
    morestring=*[d]{"},
    morestring=[s][]{\#\{}{\}},
    frame=shadowbox,
}

\newcommand{\textmachina}{\textsc{TextMachina}}

\newcommand\blfootnote[1]{%
  \begingroup
  \renewcommand\thefootnote{}\footnote{#1}%
  \addtocounter{footnote}{-1}%
  \endgroup
}

\title{\includegraphics[height=14pt]{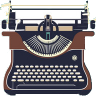}~\textmachina: Seamless Generation of Machine-Generated Text Datasets}

\author{Areg Mikael Sarvazyan$^{*}$ \and  José Ángel González$^{*}$ \and Marc Franco-Salvador \\
  Genaios, Valencia, Spain \\
  \texttt{\{areg.sarvazyan, jose.gonzalez, marc.franco\}@genaios.ai} \\
}

\begin{document}
\maketitle
\blfootnote{* Equal contribution.}

\begin{abstract}

Recent advancements in Large Language Models (LLMs) have led to high-quality Machine-Generated Text (MGT), giving rise to countless new use cases and applications.
However, easy access to LLMs is posing new challenges due to misuse. 
To address malicious usage, researchers have released datasets to effectively train models on MGT-related tasks.
Similar strategies are used to compile these datasets, but no tool currently unifies them.
In this scenario, we introduce \textmachina, a modular and extensible Python framework, designed to aid in the creation of high-quality, unbiased datasets to build robust models for MGT-related tasks such as detection, attribution, mixcase, or boundary detection.
It provides a user-friendly pipeline that abstracts away the inherent intricacies of building MGT datasets, such as LLM integrations, prompt templating, and bias mitigation.
The quality of the datasets generated by \textmachina~has been assessed in previous works, including shared tasks where more than one hundred teams trained robust MGT detectors.\footnote{Available at \url{www.github.com/Genaios/TextMachina}, and installable via \texttt{pip install text-machina}. Demo video at \url{www.youtube.com/watch?v=IcfXLRSs4Fc}}
\end{abstract}

\section{Introduction}

Recent advancements in Large Language Models (LLMs) have led to a new dawn in Machine-Generated Text (MGT).
Highly capable models such as GPT~\cite{brown2020language,ouyang2022training} and LLaMA~\cite{touvron2023llama} have established the foundations for creating massively-adopted 
services like ChatGPT,\footnote{\url{chat.openai.com}} 
fostering non-technical users to effortlessly generate high-quality, multi-style and multi-domain content and solutions \cite{eloundou2023gpts,liu2023summary}.

Unfortunately, this also risks intellectual property rights violations \cite{henderson2023foundation}, data leakage \cite{nasr2023scalable}, and malicious use that threatens the reputation of organizations and individuals \cite{kasneci2023chatgpt}, e.g., generating fake-news, polarised opinions, or smear campaigns.

\begin{figure}[!t]
\centering
\includegraphics[width=\linewidth]{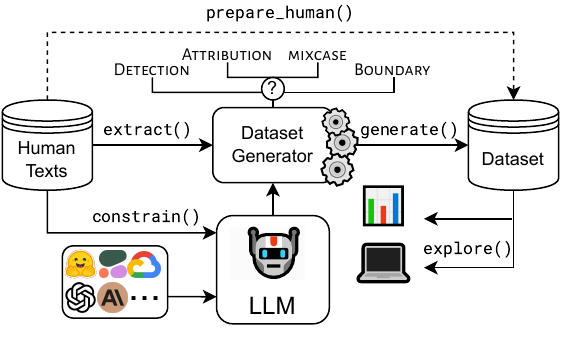}
\caption{\label{fig:overview}Overview of \textmachina's pipeline. Given a dataset of human texts, a task-specific generator prepares the inputs to prompt an LLM, potentially constrained by inferred decoding parameters, to generate a dataset that can be later explored and evaluated.}
\vspace{-1.5em}
\end{figure}

MGT detection offers competitive performance when applied to specific domains, data sources, or models \cite{bakhtin2019real}.
However, modern LLMs are generally leveraged by combining these variables in large-scale scenarios \cite{eloundou2023gpts}.
Therefore, there is a need for solutions to 1) \textbf{detect} MGT (\emph{Is this text machine generated?}), 2) \textbf{attribute} it to specific text generation models (\emph{Which model generated this text?}), 3) \textbf{detect the boundary} between machine-generated and human fragments in a text (\emph{Where does the generated content start?}), and 4) \textbf{mixcase detection} of both machine-generated and human sections in a text (\emph{Which spans of text are machine-generated?}).
Recent efforts include zero-shot approaches \cite{mitchell2023detectgpt,zellers2019defending} and supervised systems \cite{ippolito-etal-2020-automatic,uchendu-etal-2020-authorship} for document-level detection and attribution, while few works focused on mixcase tasks \cite{zhang2024llmasacoauthor}.
Some novel works have studied generalization across model families and scales \cite{clef_autextification,antoun2023text}.
Moreover, aiming to foster research and knowledge sharing in this field, different evaluation campaigns have been organized \cite{shamardina2022findings}, including multi-domain MGT detection and attribution~\cite{sarvazyan2023overview}, and multi-lingual and boundary detection settings \cite{se24t8}. 

Creating high-quality MGT datasets is essential for these initiatives.
Unfortunately, certain aspects complicate the task.
New LLMs and applications are released every day.
It is vital that detectors are robust and can generalize, meaning that datasets must offer a realistic representation of MGT tasks, where languages, domains, writing styles, and generation models are combined. 
In addition, creating MGT datasets requires prompt engineering knowledge and mechanisms so as not to incur biases, which would create unrealistic and easy-to-model datasets.

In this work, we introduce \textmachina, a modular, easy-to-use, and extensible Python framework to build datasets for MGT tasks such as detection, attribution, or boundary detection.
As depicted in Figure \ref{fig:overview}, \textmachina~provides a fully customizable pipeline that (i) abstracts away the interaction with different LLM providers, (ii) includes user-friendly information extraction, templating, constraining mechanisms to control the text generation process, and (iii) automatically prevents common biases. 
Moreover, considering the challenges that arise when working with different LLMs, our framework offers an exploration mode with utilities to aid users in understanding the dataset and task difficulty.
\textmachina~is designed such that existing dataset generation methodologies can be easily reproduced. 
Notably, an early version has successfully been used to build high-quality, unbiased datasets leveraged in shared tasks with over one hundred participating teams \cite{sarvazyan2023overview}.

\section{Challenges in MGT Dataset Generation}
Akin to human-annotated datasets, MGT dataset creation is not exempt to challenges. 
Common challenges when compiling MGT datasets include implementation overhead, model access, and controllable generation, as well as data-specific issues like bias mitigation.
\textmachina~addresses these by design.

\paragraph{Implementation overhead.} A core challenge when generating massive quantities of MGT is the general overhead that comes with needing to develop specific steps for data processing, information extraction, text generation, post-processing and dataset compilation.
To our knowledge, no tool offers a unified, re-usable, and customizable pipeline to generate MGT datasets.

\paragraph{Model access.} It is vital to be able to easily use any new and existing LLMs, considering the emergence rate of new model providers, cloud platforms, and companies that offer proprietary models, inference APIs, and model deployment services.
To generate text, one would need to examine their specifications and configure them accordingly. 

\paragraph{Controllable generation.} When generating text with multiple LLMs in many styles, one needs to explore mechanisms for controllable generation so the MGT is of expected high quality and tied to the specific domains and topics extracted from human sources. 
This requires an easy way to experiment with generation and decoding parameters (e.g., softmax temperature), as well as exploring different prompt engineering techniques.

\paragraph{Bias mitigation.}\label{ss:bias-mitigation} We refer to a bias as any artificially-introduced pattern in an MGT dataset that a trained model can exploit as a proxy to easily solve the task. 
These biases result in models that learn spurious correlations and generalize poorly in practical scenarios.
Some of them are illustrated in Table \ref{tab:bias-examples}, as a result of ineffective prompting.\footnote{See Appendix \ref{sec:bias-definitions} for a detailed description of all the biases addressed by \textmachina.}

\begin{table}[t!]
\vspace{-0.3cm}
\centering
\vspace{1em}
\resizebox{0.475\textwidth}{!}{
\begin{tabular}{ll}
\hline\noalign{\vskip 0.2cm}
\textbf{Human} & \begin{tabular}[c]{@{}l@{}}\parbox[t]{3in}{It's a \boldcolor{applegreen}{microwave} and it does microwave things in a small place. What more could you want? Also, and obvious, if you have big plates, they won't fit in this microwave.}\end{tabular} \\ \noalign{\vskip 0.2cm}\hline\noalign{\vskip 0.2cm}
\textbf{Generated}  & \begin{tabular}[c]{@{}l@{}}\parbox[t]{3in}{I am sorry, \boldcolor{carminered}{I am an AI language model, I do not have feelings and cannot effectively} \boldcolor{carminered}{write a review.} However, an example review could look like this: \\ My spouse and I recently celebrated our wedding weekend with a delightful \boldcolor{applegreen}{3-night stay} \boldcolor{applegreen}{at this enchanting hotel}. From the moment \boldcolor{babyblueeyes}{âŒ£e} stepped through the doors, we were greeted with exceptional warmth and hospitality. \boldcolor{brightlavender}{La ubicación del hotel es un oasis} \boldcolor{brightlavender}{sereno,} \boldcolor{cadmiumorange}{proporcionando:}\\ \boldcolor{cadmiumorange}{(1)} \boldcolor{brightlavender}{un lugar refugiado, alejado del bullicio y el ajetreo de la ciudad.}\\ \boldcolor{cadmiumorange}{(2)} \boldcolor{brightlavender}{una atmósfera relajante, que invita al descanso y la desconexión, ideal}…}\end{tabular} \\ \noalign{\vskip 0.2cm} \hline
\end{tabular}}
\caption{\label{tab:bias-examples}Examples of biases. Note that generated text is much longer than human text and it exhibits \boldcolor{carminered}{disclosure}, \boldcolor{applegreen}{topic}, \boldcolor{babyblueeyes}{encoding}, \boldcolor{brightlavender}{language}, and \boldcolor{cadmiumorange}{structure} biases.}
\vspace{-0.51cm}
\end{table}

\section{\textmachina}

\begin{figure}[!t]
    \centering
    \scalebox{0.96}{
\begin{lstlisting}[keywords={from, import}, numbers=none]
from text_machina import get_generator
from text_machina import Config, InputConfig, ModelConfig

config = Config(
    input=InputConfig(...),
    model=ModelConfig(...),
    generation={...},
    task_type="detection",
)

generator = get_generator(config)
dataset = generator.generate()
\end{lstlisting}
    }
    \caption{\label{fig:programmatic-usage} Using \textmachina~programmatically to generate a dataset. Ellipsis abbreviate the arguments to each type of configuration.}
    \vspace{-0.6em}
\end{figure}

\textmachina~includes all the relevant tools and functionalities to generate unbiased MGT datasets for various tasks.
It provides two core features: \texttt{explore}, for users to generate a small sample and assess generation quality and task difficulty, and \texttt{generate}, to generate complete datasets.
\textmachina~splits the dataset generation process into the following steps: (i) configuration, (ii) information extraction, (iii) MGT generation and, (iv) post-processing.
It also includes a CLI to generate and explore datasets, report metrics and statistics.
These features are offered in a modular and extensible manner so users can incorporate new model providers, extractors, task types, metrics, and more.
In the following sections we describe what these entail, and how they overcome the common challenges that arise when building MGT datasets.

\subsection{Usage}
\begin{figure}[t!]
\centering
\includegraphics[width=\linewidth, trim={1cm 1cm 1.4cm 1.2cm}, clip]{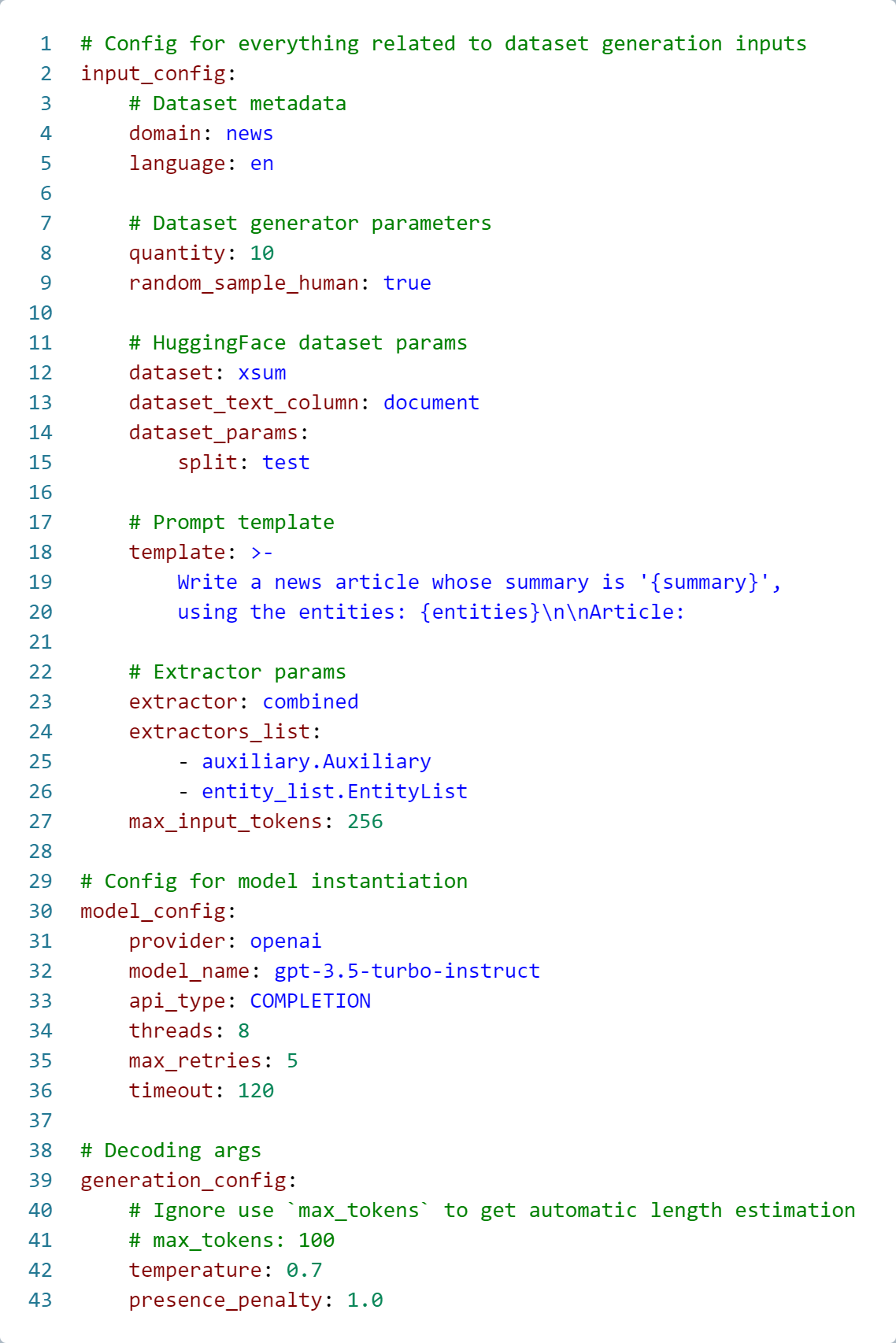}
\caption{\label{fig:config} YAML config to generate a dataset of 10 news articles, using human texts from the XSum \cite{narayan-etal-2018-dont} dataset, by prompting \texttt{gpt-3.5-turbo-instruct} using the \texttt{Combined} extractor to fill the prompt with summaries and entities.}
\end{figure}

The features offered by \textmachina~can be used through a CLI or programmatically.
In both cases, the user can specify the dataset generation parameters via customizable configurations, either through Python classes (Figure \ref{fig:programmatic-usage}) or using a YAML configuration file (Figure \ref{fig:config}).

\begin{figure*}
\centering
\includegraphics[width=\linewidth]{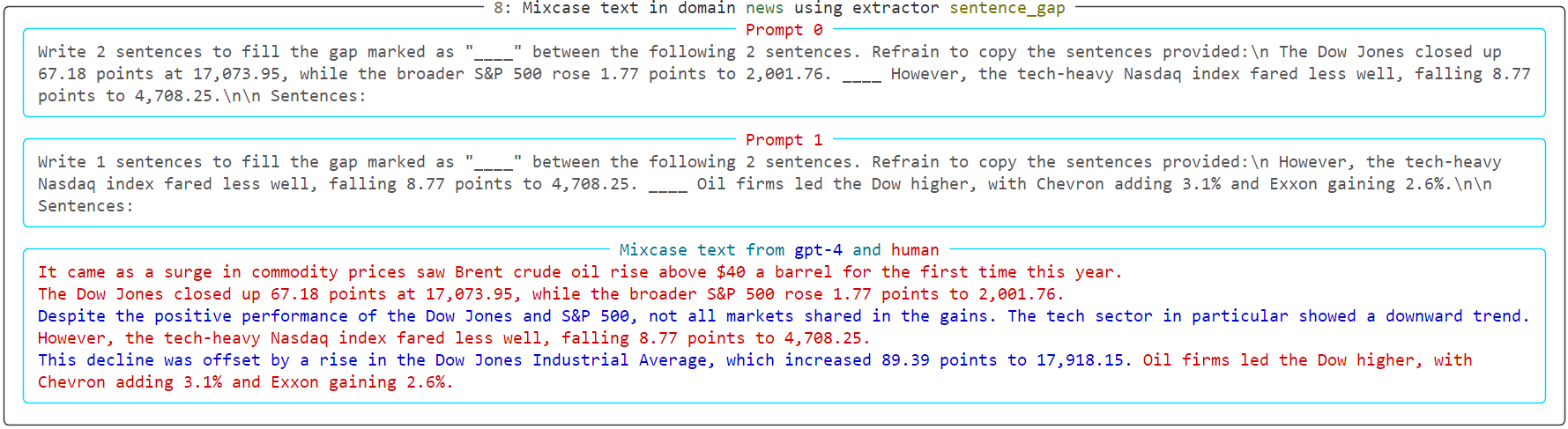}
\caption{\label{fig:cli-detection} Interactive exploration of a mixcase detection dataset.}
\end{figure*}
In the CLI, two core endpoints are available: \texttt{explore} to interactively inspect datasets and \texttt{generate} to generate complete datasets.
Both endpoints require specifying a YAML configuration file or a directory tree that contains them.\footnote{The generated dataset will be the concatenation of the datasets generated with each YAML file.}

The \texttt{explore} endpoint offers an interactive interface to check the quality and potential biases of a small dataset, which is paramount to ensure that important resources such as computation time and budget are appropriately used before generating massive datasets using \texttt{generate}. 
Figure \ref{fig:cli-detection} shows the interface to explore a mixcase detection dataset, but this interface is supported for every task type (see Figure \ref{fig:cli-boundary} in the Appendix).

The \texttt{generate} endpoint generates MGT datasets by running the \textmachina~pipeline with additional features such as caching and run systems. 
Our framework also stores the progress along the dataset generation process and is able to recover from errors, enabling users to continue interrupted runs from existing checkpoints.
\subsection{Dataset Generation}\label{ss:dataset-generators}

Different types of MGT datasets involve different generation methodologies. 
For instance, building a dataset aimed for boundary detection such as Subtask C in SemEval-2024 Task 8 \cite{se24t8} requires prepending human texts to the generations, while compiling datasets for MGT detection like AuTexTification \cite{sarvazyan2023overview} need human and generated texts to be handled independently. 
To abstract this, \textmachina~provides a \texttt{DatasetGenerator} interface to implement different generators of MGT datasets.
It accepts any human dataset in the HuggingFace Dataset format \cite{lhoest-etal-2021-datasets}, and generates a labeled dataset for a given task using the same format, including metadata such as a user-defined domain, prompt, and configuration.

Currently, \textmachina~offers generators to build datasets for the four most popular MGT-related tasks: MGT detection, model attribution, boundary detection, and mixcase detection.

\subsection{Model Providers}\label{ss:model-providers}

\textmachina~implements a way to integrate any LLM provider. 
Currently, \textmachina~integrates LLMs from Anthropic, Cohere, OpenAI, Azure OpenAI, Google Vertex AI, Amazon Bedrock, AI21, inference servers like VLLM or Triton, and any model from HuggingFace deployed either locally or remotely.\footnote{See Table \ref{tab:providers} in the Appendix for details regarding LLMs.} \textmachina~abstracts working with any provider through the \texttt{TextGenerationModel} interface, which retrieves generations from LLM providers.
\textmachina~internally handles multi-threading requests for new providers and allows an easy configuration of the decoding parameters.
It also manages recoverable errors by retrying at random increasing intervals with exponential back-off and jitter \cite{Brooker_2015}.

\begin{table*}[!ht]
    \small
    \centering
    \begin{tabular*}{\linewidth}{ll>{\em}l}
        \toprule
        \textbf{Extractor} & \textbf{Description} & \multicolumn{1}{l}{\textbf{Example Prompt Template}} \\ \toprule
        Auxiliary & Any text column  & Write a news article whose summary is {\color{codepurple}\{summary\}}, adopting the style \\
         & in the human dataset & in: {\color{codepurple}\{newspaper\}}. \\ \midrule
        Entities & Entities from & Write a fictional story with these entities: {\color{codepurple}\{entities\}}. \\
        & human texts & \\
        Nouns & Noun phrases & Write a legal document based on the following noun phrases: {\color{codepurple}\{nouns\}}.\\ 
        & from human texts & \\ \midrule
        Sentence Prefix & The first $k$ sentences  & Write a Wikipedia article starting with these two sentences: {\color{codepurple}\{sentences\}}.\\
        & of a human text  & \\ 
        Word Prefix & The first $k$ words & Write a tweet starting with these four words: {\color{codepurple}\{words\}}. \\ 
        & of a human text & \\ \midrule
        Sentence Gap & Two boundary & Write {\color{codepurple}\{n\}} sentences to fill the gap marked with "\_\_\_\_" between these 2 \\ & sentences & sentences: {\color{codepurple}\{boundaries\}} \\
        Word Gap & Two boundary &  Write {\color{codepurple}\{n\}} words to fill the gap marked with "\_\_\_\_" between these 2 \\ & word spans & word spans: {\color{codepurple}\{boundaries\}} \\ \midrule
        Sentence Masking & Masked sentences & Fill the masks, writing new sentences to be coherent with the context. \\ & to be reconstructed & Format your output according to this JSON schema: \\ & & \{\{"MASK-0": <sentence>\}, \ldots\}. Article with gaps: {\color{codepurple}\{masked\_text\}}. \\ 
        Word Masking & Masked words spans & Fill the masks, writing new word spans to be coherent with the context. \\ & to be reconstructed & Format your output according to this JSON schema: \\ & & \{\{"MASK-0": <word-span>\}, \ldots\}. Article with gaps: {\color{codepurple}\{masked\_text\}}. \\ \midrule
        Sentence Rewriting & A sentence to rewrite & Rewrite this sentence in your own words. Sentence: {\color{codepurple}\{sentence\}}. \\ \midrule
        Combined & Combines any of the & Write a text similar to this one: {\color{codepurple}\{document\}}, whose summary is {\color{codepurple}\{summary\}}, \\
         & previous extractors & using the following nouns: {\color{codepurple}\{nouns\}}; and entities: {\color{codepurple}\{entities\}}. \\
         \bottomrule
    \end{tabular*}
    \caption{\label{tab:extractors} The extractors that \textmachina~currently offers.}
\end{table*}

\subsection{Extractors}
To build MGT datasets, LLMs are asked to generate text based on a given dataset of human texts by means of a prompt. 
Examples include generating Wikipedia articles from titles, news articles from summaries, or responses to a dialog utterance.

Here, a prompt is built using information from human texts as a way to guide the LLM to generate texts about specific topics in specific styles.
Datasets such as M4 \cite{wang2023m4}, MULTITuDE \cite{macko-etal-2023-multitude}, or MGTBench \cite{he2023mgtbench} use auxiliary fields such as titles, abstracts, question-context pairs, or summaries of existing texts.
Others instruct models to paraphrase 
human texts, e.g., OpenLLMText \cite{chen-etal-2023-token} and OpenGPTText \cite{chen2023gptsentinel}.
Datasets like AuTexTification \cite{sarvazyan2023overview} and SeqXGPT \cite{wang-etal-2023-seqxgpt} leverage human text prefixes at word or sentence level.

\textmachina~covers these and more scenarios by providing an \texttt{Extractor} interface that can be used to add new extractors by implementing the \texttt{extract} and \texttt{prepare\_human} methods. 
Currently, we offer eleven types of extractors, depicted in Table \ref{tab:extractors}, that can be linked in the prompt templates to automatically fill them with information from datasets. 
These extractors can be parameterized too, e.g., to select the number of sentences to be extracted in \texttt{SentencePrefix} or the amount of words to be written between the span boundaries in \texttt{WordGap}.
\textmachina~automatically handles the prompt inputs, cleaning them to avoid breaking the prompt format, and truncating to avoid surpassing the maximum length allowed by LLMs. 
The extractors designed to generate continuations, \texttt{SentencePrefix} and \texttt{WordPrefix}, also remove the extracted prefix from the original human sources to avoid context biases (Appendix \ref{sec:bias-definitions}), thus ensuring that both the human text and the MGT are continuations of the same prefix.
Note that these extractors are also useful to prompt non-instructed LLMs, by just using the templates \textit{\color{codepurple}\{sentences\}} or \textit{\color{codepurple}\{words\}} without specifying additional instructions.

\subsection{Constrainers}

In \textmachina, we define a constrainer as any kind of logic that infers something from a human dataset and constrains the LLM's decoding parameters accordingly. 
For instance, length constrainers automatically infer the token length of the texts and return maximum and minimum number of tokens accordingly. 
One could even analyze the level of formality and diversity in human texts and infer the LLM's temperature parameter to generate texts similar to those in the dataset.
In \textmachina, users can define any kind of constrainer to automatically estimate decoding parameters. 
Currently, only length constrainers are provided by \textmachina~as an attempt to alleviate length biases, avoiding large differences in token-length between generated and human texts.

\subsection{Post-processing}

Following the best practices in the field, \textmachina~applies a set of post-processing functions by default, both to the generated and the human texts, aimed at improving the quality of any MGT dataset and preventing common biases and artifacts.
These post-processing steps are carried out in the following order to ensure they are applied correctly in a cohesive manner.

\paragraph{Language filter.} \textmachina~mitigates language bias using \texttt{fastText} \cite{joulin-etal-2017-bag} language identification models to apply language filtering on the generated texts, discarding all the texts in a language different from the human texts.

\paragraph{Fix encoding.} \textmachina~addresses encoding bias by fixing mojibake and Unicode errors with \texttt{ftfy} \cite{speer-2019-ftfy}.

\paragraph{Remove disclosure tokens and patterns.} \textmachina~ ensures that no special tokens such as \texttt{[BOS]}, \texttt{[PAD]}, or \texttt{[EOS]} appear in the generated text and removes predefined disclosure patterns like ``\textit{As a language model, ...}'' to minimize disclosure biases.

\paragraph{Remove leading and trailing whitespaces.} LLMs are prone to generating texts wrapped with whitespaces. \textmachina~removes any sequence of leading and trailing whitespaces both in the generations and the human texts.

\paragraph{Truncate.} We mitigate length biases via a truncation algorithm to ensure that every class has a similar token-length distribution. 
This algorithm first groups texts of similar token-lengths in each class and then truncates them, thus removing as little text as possible.
Moreover, all the texts containing less than 10 tokens are discarded, since typically it is not enough to correctly discriminate between generated and human.
Truncation is not applied for boundary detection.

\paragraph{Remove empty texts.} Under specific circumstances, LLMs can generate empty texts. 
\textmachina~ removes all the empty texts both from the generated and human texts.

\paragraph{Remove duplicates.} It is possible that some generations are identical to human texts in the dataset, especially when dealing with shorter texts such as tweets or chat responses. 
For boundary detection tasks, \textmachina~removes all the duplicated texts. 
For detection and attribution tasks, all the duplicated texts within and across labels are removed.
This way, we eliminate any 
label noise by making the text-label pairs disjoint.

\paragraph{Remove generation errors.} Even though \textmachina~attempts to recover from server side errors, sometimes it is not possible to generate a completion even after several retries, where \textmachina~discards it.

\subsection{Metrics}

\textmachina~provides a set of metrics to assess task difficulty and dataset quality as a way to rapidly identify failure modes. 
For instance, if the diversity of the generated texts is too low, one may wish to modify the temperature parameter or discard the LLM. 
These metrics can be applied both in the interactive phase and after generating a whole dataset. 
Five metrics are currently provided, depending on the task type: (i) MAUVE \citep{pillutla2021mauve} to measure distributional distances between classes, (ii) Repetition and Diversity \cite{su2022a} to quantify text degeneration by computing ratios between unique and total n-grams, (iii) the performance of a text classification model for document-level tasks, (iv) the performance of a token classification model for boundary and mixcase tasks, and (v) per-class averaged text perplexity.\footnote{See Table \ref{tab:metrics} in the Appendix for details regarding metrics.}

\section{Use Cases}

\textmachina~has been used to build high-quality datasets comprised of MGT in multiple languages, styles, and domains, from different LLMs and decoding strategies.

Initial versions of \textmachina~built the MGT detection and attribution datasets of the AuTexTification shared task, 
which are publicly available in HuggingFace and have been downloaded more than two thousand times.\footnote{\url{https://huggingface.co/datasets/symanto/autextification2023}} 
The AuTexTification datasets contain more than 160,000 texts in English and Spanish, from BLOOM and GPT models with different parameter scales, and five domains. 
The quality of these datasets has been assessed by more than one hundred international teams that participated in the workshop, which fostered the development of robust models for MGT detection and attribution \cite{przybyla2023ve}.

\textmachina~has also been leveraged to build the MGT detection and attribution datasets of the IberAuTexTification shared task, to be presented at the Iberian Languages Evaluation Forum (IberLEF 2024).\footnote{\url{https://sites.google.com/view/iberautextification/home}} These datasets encompass over 100,000 texts generated by state-of-the-art LLMs in languages from the Iberian Peninsula. Furthermore, \textmachina~was employed to create datasets for exploring the generalization capabilities of supervised MGT detectors across LLM's families and parameter scales \cite{clef_autextification}. 

Beyond the mentioned use cases, our framework can be used, with proper configurations, to build any existing MGT dataset in the literature. Overall, there is a large interest in building MGT datasets using \textmachina, which is reflected by more than three thousand downloads of the package in PyPI the first two months.

\section{Conclusion and Future Work}

We presented \textmachina, a modular Python framework that provides a comprehensive pipeline composed of a set of tools aimed to construct high-quality, unbiased datasets for MGT-related tasks such as detection, attribution, and boundary detection. Among these tools, \textmachina~provides (i) dataset generators to build several kinds of MGT datasets, (ii) an interface to integrate any LLM, (iii) a set of extractors to fill prompt templates with information from human text datasets, (iv) constrainers to automatically infer LLM decoding hyperparameters, (v) post-processing functions, and (vi) a user-friendly CLI to generate and explore datasets. 
We have also shown common biases that could be artificially introduced in MGT datasets and how we can help the users to prevent them. \textmachina~has proved its potential as a reliable tool to create MGT datasets used in shared tasks with more than one hundred participating teams.

Future work includes adding new capabilities to \textmachina, e.g., supporting additional LLM providers and addressing any new challenge in the field.

\section*{Acknowledgements}

We thank Ian Borrego, Angelo Basile, Mara Chinea, and Stuart Winter-Tear for their support and comments.

\section*{Ethics Statement}
Large Language Models raise concerns regarding their potential misuse for generating deceptive or harmful content. \textmachina~ is developed with the ethical intent of facilitating the development of MGT detectors, enhancing their robustness and reliability, and promoting a responsible and ethical use of LLMs.

\bibliography{anthology,custom}

\begin{thebibliography}{35}
\expandafter\ifx\csname natexlab\endcsname\relax\def\natexlab#1{#1}\fi

\bibitem[{Antoun et~al.(2023)Antoun, Sagot, and Seddah}]{antoun2023text}
Wissam Antoun, Beno{\^\i}t Sagot, and Djam{\'e} Seddah. 2023.
\newblock From text to source: Results in detecting large language model-generated content.
\newblock \emph{arXiv preprint arXiv:2309.13322}.

\bibitem[{Bakhtin et~al.(2019)Bakhtin, Gross, Ott, Deng, Ranzato, and Szlam}]{bakhtin2019real}
Anton Bakhtin, Sam Gross, Myle Ott, Yuntian Deng, Marc'Aurelio Ranzato, and Arthur Szlam. 2019.
\newblock Real or fake? learning to discriminate machine from human generated text.
\newblock \emph{arXiv preprint arXiv:1906.03351}.

\bibitem[{Brooker(2015)}]{Brooker_2015}
Marc Brooker. 2015.
\newblock \href {https://aws.amazon.com/blogs/architecture/exponential-backoff-and-jitter/} {Exponential backoff and jitter}.

\bibitem[{Brown et~al.(2020)Brown, Mann, Ryder, Subbiah, Kaplan, Dhariwal, Neelakantan, Shyam, Sastry, Askell et~al.}]{brown2020language}
Tom Brown, Benjamin Mann, Nick Ryder, Melanie Subbiah, Jared~D Kaplan, Prafulla Dhariwal, Arvind Neelakantan, Pranav Shyam, Girish Sastry, Amanda Askell, et~al. 2020.
\newblock Language models are few-shot learners.
\newblock \emph{Advances in neural information processing systems}, 33:1877--1901.

\bibitem[{Chalkidis et~al.(2019)Chalkidis, Fergadiotis, Malakasiotis, and Androutsopoulos}]{chalkidis-etal-2019-large}
Ilias Chalkidis, Emmanouil Fergadiotis, Prodromos Malakasiotis, and Ion Androutsopoulos. 2019.
\newblock \href {https://doi.org/10.18653/v1/P19-1636} {Large-scale multi-label text classification on {EU} legislation}.
\newblock In \emph{Proceedings of the 57th Annual Meeting of the Association for Computational Linguistics}, pages 6314--6322, Florence, Italy. Association for Computational Linguistics.

\bibitem[{Chen et~al.(2023{\natexlab{a}})Chen, Kang, Zhai, Li, Singh, and Raj}]{chen-etal-2023-token}
Yutian Chen, Hao Kang, Vivian Zhai, Liangze Li, Rita Singh, and Bhiksha Raj. 2023{\natexlab{a}}.
\newblock \href {https://doi.org/10.18653/v1/2023.emnlp-main.810} {Token prediction as implicit classification to identify {LLM}-generated text}.
\newblock In \emph{Proceedings of the 2023 Conference on Empirical Methods in Natural Language Processing}, pages 13112--13120, Singapore. Association for Computational Linguistics.

\bibitem[{Chen et~al.(2023{\natexlab{b}})Chen, Kang, Zhai, Li, Singh, and Ramakrishnan}]{chen2023gptsentinel}
Yutian Chen, Hao Kang, Vivian Zhai, Liangze Li, Rita Singh, and Bhiksha Ramakrishnan. 2023{\natexlab{b}}.
\newblock Gpt-sentinel: Distinguishing human and chatgpt generated content.
\newblock \emph{arXiv preprint arXiv:2305.07969}.

\bibitem[{Eloundou et~al.(2023)Eloundou, Manning, Mishkin, and Rock}]{eloundou2023gpts}
Tyna Eloundou, Sam Manning, Pamela Mishkin, and Daniel Rock. 2023.
\newblock Gpts are gpts: An early look at the labor market impact potential of large language models.
\newblock \emph{arXiv preprint arXiv:2303.10130}.

\bibitem[{He et~al.(2023)He, Shen, Chen, Backes, and Zhang}]{he2023mgtbench}
Xinlei He, Xinyue Shen, Zeyuan Chen, Michael Backes, and Yang Zhang. 2023.
\newblock Mgtbench: Benchmarking machine-generated text detection.
\newblock \emph{arXiv preprint arXiv:2303.14822}.

\bibitem[{Henderson et~al.(2023)Henderson, Li, Jurafsky, Hashimoto, Lemley, and Liang}]{henderson2023foundation}
Peter Henderson, Xuechen Li, Dan Jurafsky, Tatsunori Hashimoto, Mark~A Lemley, and Percy Liang. 2023.
\newblock Foundation models and fair use.
\newblock \emph{arXiv preprint arXiv:2303.15715}.

\bibitem[{Ippolito et~al.(2020)Ippolito, Duckworth, Callison-Burch, and Eck}]{ippolito-etal-2020-automatic}
Daphne Ippolito, Daniel Duckworth, Chris Callison-Burch, and Douglas Eck. 2020.
\newblock \href {https://doi.org/10.18653/v1/2020.acl-main.164} {Automatic detection of generated text is easiest when humans are fooled}.
\newblock In \emph{Proceedings of the 58th Annual Meeting of the Association for Computational Linguistics}, pages 1808--1822, Online. Association for Computational Linguistics.

\bibitem[{Joulin et~al.(2017)Joulin, Grave, Bojanowski, and Mikolov}]{joulin-etal-2017-bag}
Armand Joulin, Edouard Grave, Piotr Bojanowski, and Tomas Mikolov. 2017.
\newblock \href {https://aclanthology.org/E17-2068} {Bag of tricks for efficient text classification}.
\newblock In \emph{Proceedings of the 15th Conference of the {E}uropean Chapter of the Association for Computational Linguistics: Volume 2, Short Papers}, pages 427--431, Valencia, Spain. Association for Computational Linguistics.

\bibitem[{Kasneci et~al.(2023)Kasneci, Se{\ss}ler, K{\"u}chemann, Bannert, Dementieva, Fischer et~al.}]{kasneci2023chatgpt}
Enkelejda Kasneci, Kathrin Se{\ss}ler, Stefan K{\"u}chemann, Maria Bannert, Daryna Dementieva, Frank Fischer, et~al. 2023.
\newblock Chatgpt for good? on opportunities and challenges of large language models for education.
\newblock \emph{Learning and Individual Differences}, page 102274.

\bibitem[{Lhoest et~al.(2021)Lhoest, Villanova~del Moral, Jernite, Thakur, von Platen, Patil, Chaumond, Drame, Plu, Tunstall, Davison, {\v{S}}a{\v{s}}ko, Chhablani, Malik, Brandeis, Le~Scao, Sanh, Xu, Patry, McMillan-Major, Schmid, Gugger, Delangue, Matussi{\`e}re, Debut, Bekman, Cistac, Goehringer, Mustar, Lagunas, Rush, and Wolf}]{lhoest-etal-2021-datasets}
Quentin Lhoest, Albert Villanova~del Moral, Yacine Jernite, Abhishek Thakur, Patrick von Platen, Suraj Patil, Julien Chaumond, Mariama Drame, Julien Plu, Lewis Tunstall, Joe Davison, Mario {\v{S}}a{\v{s}}ko, Gunjan Chhablani, Bhavitvya Malik, Simon Brandeis, Teven Le~Scao, Victor Sanh, Canwen Xu, Nicolas Patry, Angelina McMillan-Major, Philipp Schmid, Sylvain Gugger, Cl{\'e}ment Delangue, Th{\'e}o Matussi{\`e}re, Lysandre Debut, Stas Bekman, Pierric Cistac, Thibault Goehringer, Victor Mustar, Fran{\c{c}}ois Lagunas, Alexander Rush, and Thomas Wolf. 2021.
\newblock \href {https://doi.org/10.18653/v1/2021.emnlp-demo.21} {Datasets: A community library for natural language processing}.
\newblock In \emph{Proceedings of the 2021 Conference on Empirical Methods in Natural Language Processing: System Demonstrations}, pages 175--184, Online and Punta Cana, Dominican Republic. Association for Computational Linguistics.

\bibitem[{Liu et~al.(2023)Liu, Han, Ma, Zhang, Yang, Tian, He, Li, He, Liu et~al.}]{liu2023summary}
Yiheng Liu, Tianle Han, Siyuan Ma, Jiayue Zhang, Yuanyuan Yang, Jiaming Tian, Hao He, Antong Li, Mengshen He, Zhengliang Liu, et~al. 2023.
\newblock Summary of chatgpt/gpt-4 research and perspective towards the future of large language models.
\newblock \emph{arXiv preprint arXiv:2304.01852}.

\bibitem[{Macko et~al.(2023)Macko, Moro, Uchendu, Lucas, Yamashita, Pikuliak, Srba, Le, Lee, Simko, and Bielikova}]{macko-etal-2023-multitude}
Dominik Macko, Robert Moro, Adaku Uchendu, Jason Lucas, Michiharu Yamashita, Mat{\'u}{\v{s}} Pikuliak, Ivan Srba, Thai Le, Dongwon Lee, Jakub Simko, and Maria Bielikova. 2023.
\newblock \href {https://aclanthology.org/2023.emnlp-main.616} {{MULTIT}u{DE}: Large-scale multilingual machine-generated text detection benchmark}.
\newblock In \emph{Proceedings of the 2023 Conference on Empirical Methods in Natural Language Processing}, pages 9960--9987, Singapore. Association for Computational Linguistics.

\bibitem[{Mitchell et~al.(2023)Mitchell, Lee, Khazatsky, Manning, and Finn}]{mitchell2023detectgpt}
Eric Mitchell, Yoonho Lee, Alexander Khazatsky, Christopher~D Manning, and Chelsea Finn. 2023.
\newblock Detectgpt: Zero-shot machine-generated text detection using probability curvature.
\newblock \emph{International Conference on Machine Learning}.

\bibitem[{Nakayama(2018)}]{seqeval}
Hiroki Nakayama. 2018.
\newblock \href {https://github.com/chakki-works/seqeval} {{seqeval}: A python framework for sequence labeling evaluation}.
\newblock Software available from https://github.com/chakki-works/seqeval.

\bibitem[{Narayan et~al.(2018)Narayan, Cohen, and Lapata}]{narayan-etal-2018-dont}
Shashi Narayan, Shay~B. Cohen, and Mirella Lapata. 2018.
\newblock \href {https://doi.org/10.18653/v1/D18-1206} {Don{'}t give me the details, just the summary! topic-aware convolutional neural networks for extreme summarization}.
\newblock In \emph{Proceedings of the 2018 Conference on Empirical Methods in Natural Language Processing}, pages 1797--1807, Brussels, Belgium. Association for Computational Linguistics.

\bibitem[{Nasr et~al.(2023)Nasr, Carlini, Hayase, Jagielski, Cooper, Ippolito, Choquette-Choo, Wallace, Tram{\`e}r, and Lee}]{nasr2023scalable}
Milad Nasr, Nicholas Carlini, Jonathan Hayase, Matthew Jagielski, A~Feder Cooper, Daphne Ippolito, Christopher~A Choquette-Choo, Eric Wallace, Florian Tram{\`e}r, and Katherine Lee. 2023.
\newblock Scalable extraction of training data from (production) language models.
\newblock \emph{arXiv preprint arXiv:2311.17035}.

\bibitem[{Ouyang et~al.(2022)Ouyang, Wu, Jiang, Almeida, Wainwright, Mishkin, Zhang, Agarwal, Slama, Gray, Schulman, Hilton, Kelton, Miller, Simens, Askell, Welinder, Christiano, Leike, and Lowe}]{ouyang2022training}
Long Ouyang, Jeffrey Wu, Xu~Jiang, Diogo Almeida, Carroll Wainwright, Pamela Mishkin, Chong Zhang, Sandhini Agarwal, Katarina Slama, Alex Gray, John Schulman, Jacob Hilton, Fraser Kelton, Luke Miller, Maddie Simens, Amanda Askell, Peter Welinder, Paul Christiano, Jan Leike, and Ryan Lowe. 2022.
\newblock \href {https://openreview.net/forum?id=TG8KACxEON} {Training language models to follow instructions with human feedback}.
\newblock In \emph{Advances in Neural Information Processing Systems}.

\bibitem[{Pillutla et~al.(2021)Pillutla, Swayamdipta, Zellers, Thickstun, Welleck, Choi, and Harchaoui}]{pillutla2021mauve}
Krishna Pillutla, Swabha Swayamdipta, Rowan Zellers, John Thickstun, Sean Welleck, Yejin Choi, and Zaid Harchaoui. 2021.
\newblock \href {https://openreview.net/forum?id=Tqx7nJp7PR} {{MAUVE}: Measuring the gap between neural text and human text using divergence frontiers}.
\newblock In \emph{Advances in Neural Information Processing Systems}.

\bibitem[{Przyby{\l}a et~al.(2023)Przyby{\l}a, Duran-Silva, and Egea-G{\'o}mez}]{przybyla2023ve}
Piotr Przyby{\l}a, Nicolau Duran-Silva, and Santiago Egea-G{\'o}mez. 2023.
\newblock I’ve seen things you machines wouldn’t believe: Measuring content predictability to identify automatically-generated text.
\newblock In \emph{Proceedings of the Iberian Languages Evaluation Forum (IberLEF 2023). CEUR Workshop Proceedings, CEUR-WS, Ja{\'e}n, Spain}.

\bibitem[{Sarvazyan et~al.(2023{\natexlab{a}})Sarvazyan, Gonz{\'a}lez, Franco-Salvador, Rangel, Chulvi, and Rosso}]{sarvazyan2023overview}
Areg~Mikael Sarvazyan, Jos{\'e}~{\'A}ngel Gonz{\'a}lez, Marc Franco-Salvador, Francisco Rangel, Berta Chulvi, and Paolo Rosso. 2023{\natexlab{a}}.
\newblock Overview of autextification at iberlef 2023: Detection and attribution of machine-generated text in multiple domains.
\newblock \emph{Sociedad Española de Procesamiento del Languaje Natural (SEPLN)}, 71:275--288.

\bibitem[{Sarvazyan et~al.(2023{\natexlab{b}})Sarvazyan, González, Franco-Salvador, and Rosso}]{clef_autextification}
Areg~Mikael Sarvazyan, José~\'{A}ngel González, Marc Franco-Salvador, and Paolo Rosso. 2023{\natexlab{b}}.
\newblock Supervised machine-generated text detectors: Family and scale matters.
\newblock In \emph{Information Access Evaluation meets Multilinguality, Multimodality, and Visualization}. Springer International Publishing.

\bibitem[{Shamardina et~al.(2022)Shamardina, Mikhailov, Chernianskii, Fenogenova, Saidov, Valeeva, Shavrina, Smurov, Tutubalina, and Artemova}]{shamardina2022findings}
Tatiana Shamardina, Vladislav Mikhailov, Daniil Chernianskii, Alena Fenogenova, Marat Saidov, Anastasiya Valeeva, Tatiana Shavrina, Ivan Smurov, Elena Tutubalina, and Ekaterina Artemova. 2022.
\newblock Findings of the the ruatd shared task 2022 on artificial text detection in russian.
\newblock \emph{arXiv preprint arXiv:2206.01583}.

\bibitem[{Speer(2019)}]{speer-2019-ftfy}
Robyn Speer. 2019.
\newblock \href {https://doi.org/10.5281/zenodo.2591652} {ftfy}.
\newblock Zenodo.
\newblock Version 5.5.

\bibitem[{Su et~al.(2022)Su, Lan, Wang, Yogatama, Kong, and Collier}]{su2022a}
Yixuan Su, Tian Lan, Yan Wang, Dani Yogatama, Lingpeng Kong, and Nigel Collier. 2022.
\newblock \href {https://openreview.net/forum?id=V88BafmH9Pj} {A contrastive framework for neural text generation}.
\newblock In \emph{Advances in Neural Information Processing Systems}.

\bibitem[{Touvron et~al.(2023)Touvron, Martin, Stone, Albert, Almahairi, Babaei, Bashlykov, Batra, Bhargava, Bhosale et~al.}]{touvron2023llama}
Hugo Touvron, Louis Martin, Kevin Stone, Peter Albert, Amjad Almahairi, Yasmine Babaei, Nikolay Bashlykov, Soumya Batra, Prajjwal Bhargava, Shruti Bhosale, et~al. 2023.
\newblock Llama 2: Open foundation and fine-tuned chat models.
\newblock \emph{arXiv preprint arXiv:2307.09288}.

\bibitem[{Uchendu et~al.(2020)Uchendu, Le, Shu, and Lee}]{uchendu-etal-2020-authorship}
Adaku Uchendu, Thai Le, Kai Shu, and Dongwon Lee. 2020.
\newblock \href {https://doi.org/10.18653/v1/2020.emnlp-main.673} {Authorship attribution for neural text generation}.
\newblock In \emph{Proceedings of the 2020 Conference on Empirical Methods in Natural Language Processing (EMNLP)}, pages 8384--8395, Online. Association for Computational Linguistics.

\bibitem[{Wang et~al.(2023{\natexlab{a}})Wang, Li, Ren, Jiang, Zhang, and Qiu}]{wang-etal-2023-seqxgpt}
Pengyu Wang, Linyang Li, Ke~Ren, Botian Jiang, Dong Zhang, and Xipeng Qiu. 2023{\natexlab{a}}.
\newblock \href {https://aclanthology.org/2023.emnlp-main.73} {{S}eq{XGPT}: Sentence-level {AI}-generated text detection}.
\newblock In \emph{Proceedings of the 2023 Conference on Empirical Methods in Natural Language Processing}, pages 1144--1156, Singapore. Association for Computational Linguistics.

\bibitem[{Wang et~al.(2024)Wang, Aji, Shelmanov, Whitehouse, Ivanov, Mansurov, Su, Mahmoud, Afzal, and Nakov}]{se24t8}
Yuxia Wang, Alham~Fikri Aji, Artem Shelmanov, Chenxi Whitehouse, Petar Ivanov, Jonibek Mansurov, Jinyan Su, Tarek Mahmoud, Osama~Mohammed Afzal, and Preslav Nakov. 2024.
\newblock Semeval-2024 task 8: Multigenerator, multidomain, and multilingual black-box machine-generated text detection.
\newblock \url{https://github.com/mbzuai-nlp/SemEval2024-task8}.

\bibitem[{Wang et~al.(2023{\natexlab{b}})Wang, Mansurov, Ivanov, Su, Shelmanov, Tsvigun, Whitehouse, Afzal, Mahmoud, Aji et~al.}]{wang2023m4}
Yuxia Wang, Jonibek Mansurov, Petar Ivanov, Jinyan Su, Artem Shelmanov, Akim Tsvigun, Chenxi Whitehouse, Osama~Mohammed Afzal, Tarek Mahmoud, Alham~Fikri Aji, et~al. 2023{\natexlab{b}}.
\newblock M4: Multi-generator, multi-domain, and multi-lingual black-box machine-generated text detection.
\newblock \emph{arXiv preprint arXiv:2305.14902}.

\bibitem[{Zellers et~al.(2019)Zellers, Holtzman, Rashkin, Bisk, Farhadi, Roesner, and Choi}]{zellers2019defending}
Rowan Zellers, Ari Holtzman, Hannah Rashkin, Yonatan Bisk, Ali Farhadi, Franziska Roesner, and Yejin Choi. 2019.
\newblock Defending against neural fake news.
\newblock \emph{Advances in neural information processing systems}, 32.

\bibitem[{Zhang et~al.(2024)Zhang, Gao, Chen, Huang, Huang, Sun, Zhang, Li, Fu, Wan, and Sun}]{zhang2024llmasacoauthor}
Qihui Zhang, Chujie Gao, Dongping Chen, Yue Huang, Yixin Huang, Zhenyang Sun, Shilin Zhang, Weiye Li, Zhengyan Fu, Yao Wan, and Lichao Sun. 2024.
\newblock \href {http://arxiv.org/abs/2401.05952} {Llm-as-a-coauthor: Can mixed human-written and machine-generated text be detected?}

\end{thebibliography}
\bibliographystyle{acl_natbib}

\appendix
\section{Biases considered in \textmachina}\label{sec:bias-definitions}

The biases discussed in section \ref{ss:bias-mitigation} with a human and a generated text can arise in the following cases. \textmachina~addresses these biases by design.

\paragraph{Length bias.} Arises when human and generated texts have very different length distributions. 

\paragraph{Topic bias.} Emerges if human and generated texts differ greatly in  topic content.
A prominent example is when both texts are in the reviews domain but human texts always review hotels, whereas generated texts review cars, books,
or electronics. 

\paragraph{Structure bias.} Occurs in domains that have a strongly-structured text which can be hard to reproduce by LLMs, e.g., legal documents as those from EUR-Lex \cite{chalkidis-etal-2019-large}. 

\paragraph{Disclosure bias.} Appears when LLMs explicitly disclose that they are LLMs through patterns such as ``\textit{As an AI, ...}'' or tokens like \texttt{[BOS]} or \texttt{[EOS]}. 

\paragraph{Encoding bias.} Arises when LLMs systematically generate symbols in a different encoding to human texts. 
Especially prominent in multilingual models or emoji-rich domains.

\paragraph{Language bias.} Emerges when LLMs predominantly generate text in a language different from that of the human texts. 

\paragraph{Context bias.} Occurs specifically while generating continuations given a prefix, when human and generated texts differ in how they treat the prefixes. 
For instance, given a human text ``\textit{I have two coins of 1€, so I have 2€}'', an LLM prompted with ``\textit{I have two coins of 1€, so I have}'' would just output ``\textit{2€}''. 
Hence, generated and human texts must either both include or exclude the prefixes.

\begin{table*}[!ht]
\footnotesize
\centering
\begin{tabular}{lll}
\toprule
\textbf{Provider} & \textbf{Models} & \textbf{URL} \\ \midrule
Anthropic & Claude-3, Claude-2, ... & \url{https://www.anthropic.com} \\
Cohere & Command, Command-Light, ... & \url{https://cohere.com} \\
OpenAI & GPT-3.5-turbo, GPT-4, ... & \url{https://openai.com} \\
Azure OpenAI & GPT-3.5-turbo, GPT-4, ... & \url{https://azure.microsoft.com} \\
Vertex AI & PaLM2, Gemini, ... & \url{https://cloud.google.com/vertex-ai} \\
Amazon Bedrock & Titan, Claude, LLama-2, ... & \url{https://aws.amazon.com/bedrock/} \\
AI21 & Jurassic-2-Ultra, Jurassic-2-Mid, ... & \url{https://www.ai21.com/} \\
HuggingFace & Mixtral, Llama-2, ... & \url{https://huggingface.co} \\
HF Inference Endpoints & Mixtral, Llama-2, ... & \url{https://huggingface.co/inference-endpoints} \\
HF Inference API & Mixtral, Llama-2, ... & \url{https://huggingface.co/inference-api} \\ 
VLLM & Any locally deployed model & \url{https://github.com/vllm-project/vllm} \\
Triton & Any locally deployed model & \url{https://github.com/triton-inference-server} \\
\bottomrule
\end{tabular}
\caption{\label{tab:providers} Model providers supported by \textmachina.}
\end{table*}

\section{Using the CLI}
\label{sec:using_cli}
\begin{figure*}[h!]
\centering
\includegraphics[width=\linewidth]{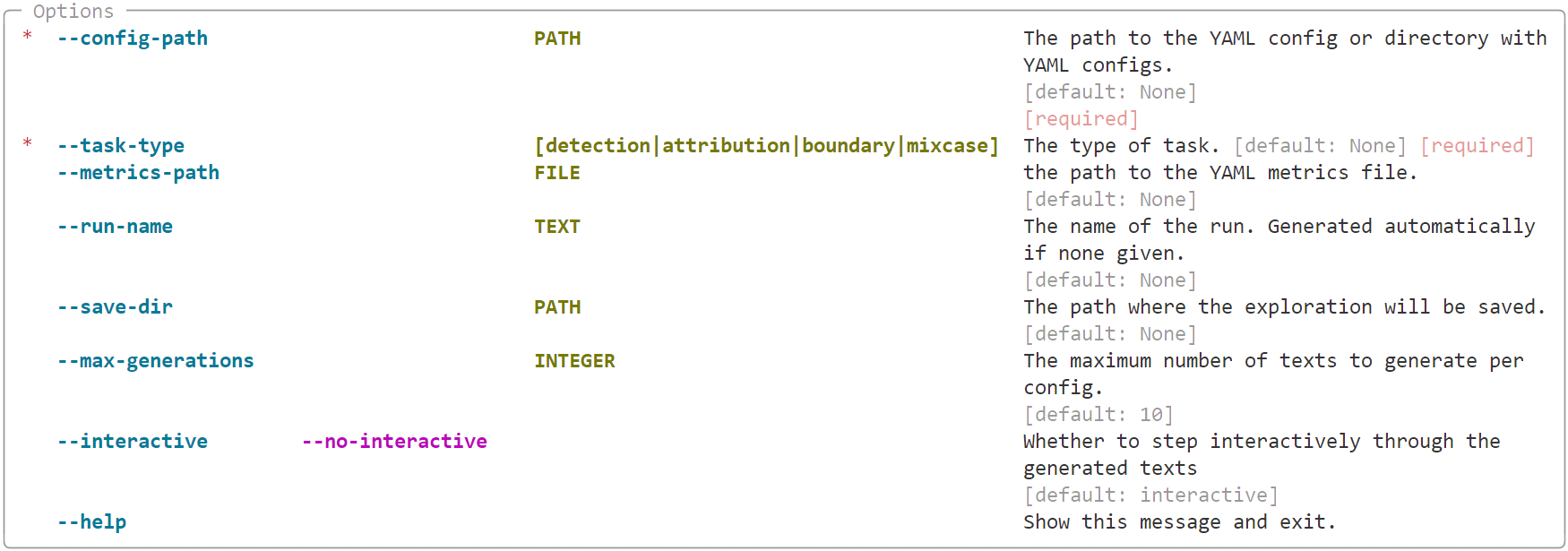}
\caption{\label{fig:explore-usage} Arguments to the \texttt{explore} endpoint.}
\end{figure*}

\begin{figure*}[h!]
\centering
\includegraphics[width=\linewidth]{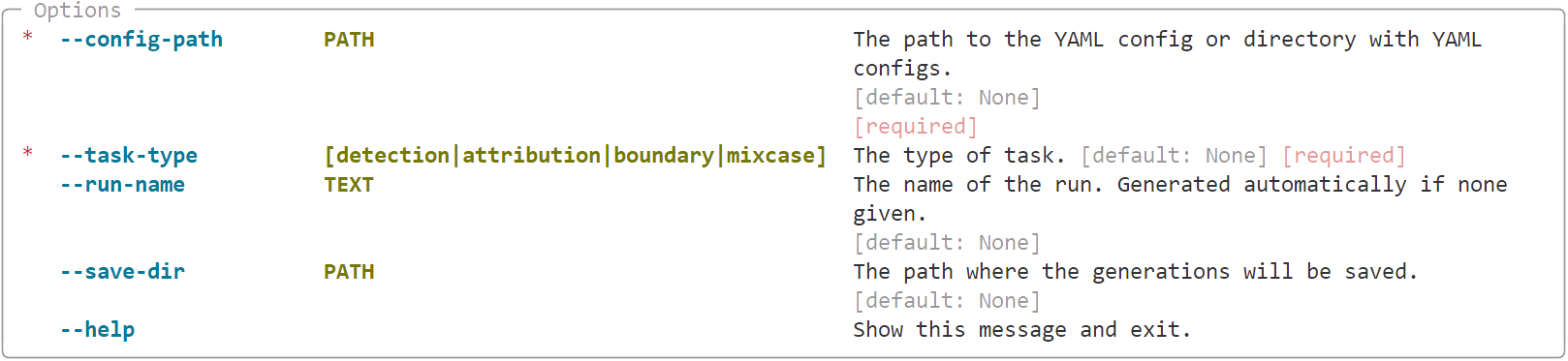}
\caption{\label{fig:generate-usage} Arguments to the \texttt{generate} endpoint.}
\end{figure*}

After installing \textmachina~as a package, users can run the \texttt{explore} and \texttt{generate} endpoints in the shell as
\lstinline{text-machina [endpoint] [args]}.
Otherwise, the CLI of \textmachina~can also be run as \lstinline{python -m text_machina.cli [endpoint] [args]} in the root folder. 
Figures \ref{fig:explore-usage} and \ref{fig:generate-usage} show the arguments of each endpoint.
We show a set of common use cases in Table \ref{tab:use-cases} and videos to illustrate how the \href{https://streamable.com/0sr5ky}{\texttt{generate}} and \href{https://streamable.com/vje6zr}{\texttt{explore}} and  endpoints work.

\begin{table*}[h!]
\small
\centering
\begin{tabular}{l>{\fontsize{8}{8}\ttfamily}ll>{\fontsize{8}{8}\ttfamily}l}
\toprule

\textbf{Use case} & \multicolumn{1}{l}{\textbf{Command}} & \textbf{Use case} & \multicolumn{1}{l}{\textbf{Command}} \\ \toprule

\begin{tabular}[c]{@{}l@{}}Generate a dataset \\ for MGT detection\end{tabular} & \begin{tabular}[c]{@{}l@{}}text-machina generate \textbackslash\\-{}-config-path config.yaml \textbackslash\\-{}-task-type detection\end{tabular} &

\begin{tabular}[c]{@{}l@{}}Generate a dataset \\ for boundary detection\end{tabular} & \begin{tabular}[c]{@{}l@{}}text-machina generate \textbackslash\\-{}-config-path config.yaml \textbackslash\\-{}-task-type boundary\end{tabular} \\ \midrule

\begin{tabular}[c]{@{}l@{}} Generate a dataset\\ for MGT attribution \end{tabular}& \begin{tabular}[c]{@{}l@{}}text-machina generate \textbackslash\\-{}-config-path config.yaml \textbackslash\\-{}-task-type attribution\end{tabular} &

\begin{tabular}[c]{@{}l@{}}Generate a dataset \\ for mixcase detection\end{tabular} & \begin{tabular}[c]{@{}l@{}}text-machina generate \textbackslash\\ -{}-config-path config.yaml \textbackslash\\-{}-task-type mixcase\end{tabular} \\ \midrule

\begin{tabular}[c]{@{}l@{}}Generate a dataset \\for MGT detection \\ using config files \\in a directory tree\end{tabular} & \begin{tabular}[c]{@{}l@{}}text-machina generate \textbackslash\\-{}-config-path configs/ \textbackslash\\-{}-task-type detection\end{tabular} &

\begin{tabular}[c]{@{}l@{}}Continue generating \\a dataset for MGT \\detection from an \\interrupted process\end{tabular} & \begin{tabular}[c]{@{}l@{}}text-machina generate \textbackslash\\-{}-config-path config.yaml \textbackslash\\-{}-task-type detection \textbackslash\\-{}-run-name greedy-bear\end{tabular} \\ \midrule

\begin{tabular}[c]{@{}l@{}}Explore a dataset  \\ of 10 samples for \\ MGT detection \\ and compute metrics\end{tabular} & \begin{tabular}[c]{@{}l@{}}text-machina explore \textbackslash\\ -{}-config-path config.yaml \textbackslash\\-{}-task-type detection \textbackslash\\-{}-max-generations 10 \textbackslash\\-{}-metrics-path metrics.yaml\end{tabular} &

\begin{tabular}[c]{@{}l@{}}Explore an existing \\dataset for MGT detection \\and compute metrics\end{tabular} & \begin{tabular}[c]{@{}l@{}}text-machina explore \textbackslash\\-{}-config-path config.yaml \textbackslash\\-{}-task-type detection \textbackslash\\-{}-run-name greedy-bear \textbackslash\\-{}-metrics-path metrics.yaml\end{tabular} \\
\bottomrule
\end{tabular}
\caption{\label{tab:use-cases} Common use cases and commands in \textmachina.}
\end{table*}

\begin{table*}[h!]
    \small
    \centering
    \begin{tabular}{lll}
    \toprule 
    \textbf{Metric}  & \textbf{Tasks} & \textbf{Description}  \\ \toprule
    MAUVE \cite{pillutla2021mauve} & \begin{tabular}[c]{@{}l@{}}detection\\attribution\end{tabular} & \begin{tabular}[c]{@{}l@{}}Measure divergences between all the classes using quantized\\  embeddings of a small pre-trained language model.\end{tabular} \\ \midrule
    Text Perplexity & \begin{tabular}[c]{@{}l@{}}detection\\attribution\\boundary\end{tabular} & \begin{tabular}[c]{@{}l@{}}Average per-class perplexity.\\ For boundary it treats human and generated segments separately. \end{tabular}\\ \midrule
    \begin{tabular}[c]{@{}l@{}}Repetition \& Diversity\\ \cite{su2022a}\end{tabular} & \begin{tabular}[c]{@{}l@{}}detection\\attribution\\boundary\end{tabular} & \begin{tabular}[c]{@{}l@{}}$\text{rep}_n(y) = 100 \times (1.0 - \frac{|\text{unique } n\text{-grams}(y)|}{|\text{total } n\text{-grams}(y)|}), n \in \{2,3,4\}$\\ $\text{diversity}(y) = \prod_{n=2}^4 (1.0 - \frac{\text{rep}_n(y)}{100})$ \\ For boundary it treats human and generated segments separately.\end{tabular}\\ \midrule
    \begin{tabular}[c]{@{}l@{}}Classification model\end{tabular} & \begin{tabular}[c]{@{}l@{}}detection\\attribution\\boundary\end{tabular} & \begin{tabular}[c]{@{}l@{}}For detection and attribution it is a logistic regression with\\ bag-of-word and bag-of-character features. \\ For boundary, it predicts the point with maximal \\difference in readability scores between a prefix and suffix.\end{tabular}\\ \midrule
    \begin{tabular}[c]{@{}l@{}}Token classification model\end{tabular} & \begin{tabular}[c]{@{}l@{}}boundary\\mixcase\end{tabular} & \begin{tabular}[c]{@{}l@{}}A HuggingFace model for token classification. The tokens of human \\ fragments are labeled as \textit{human} and the generated ones as \textit{generated}. \\ Precision, recall, and $F_1$ between predictions and references are \\ computed using the \textit{seqeval} framework \cite{seqeval}\end{tabular}\\
    \bottomrule
    \end{tabular}
    \caption{\label{tab:metrics}Detailed description of the metrics offered by \textmachina.}
\end{table*}

\section{Generating a Boundary Detection Dataset}
\begin{figure*}[!ht]
\centering
\includegraphics[scale=0.36, clip]{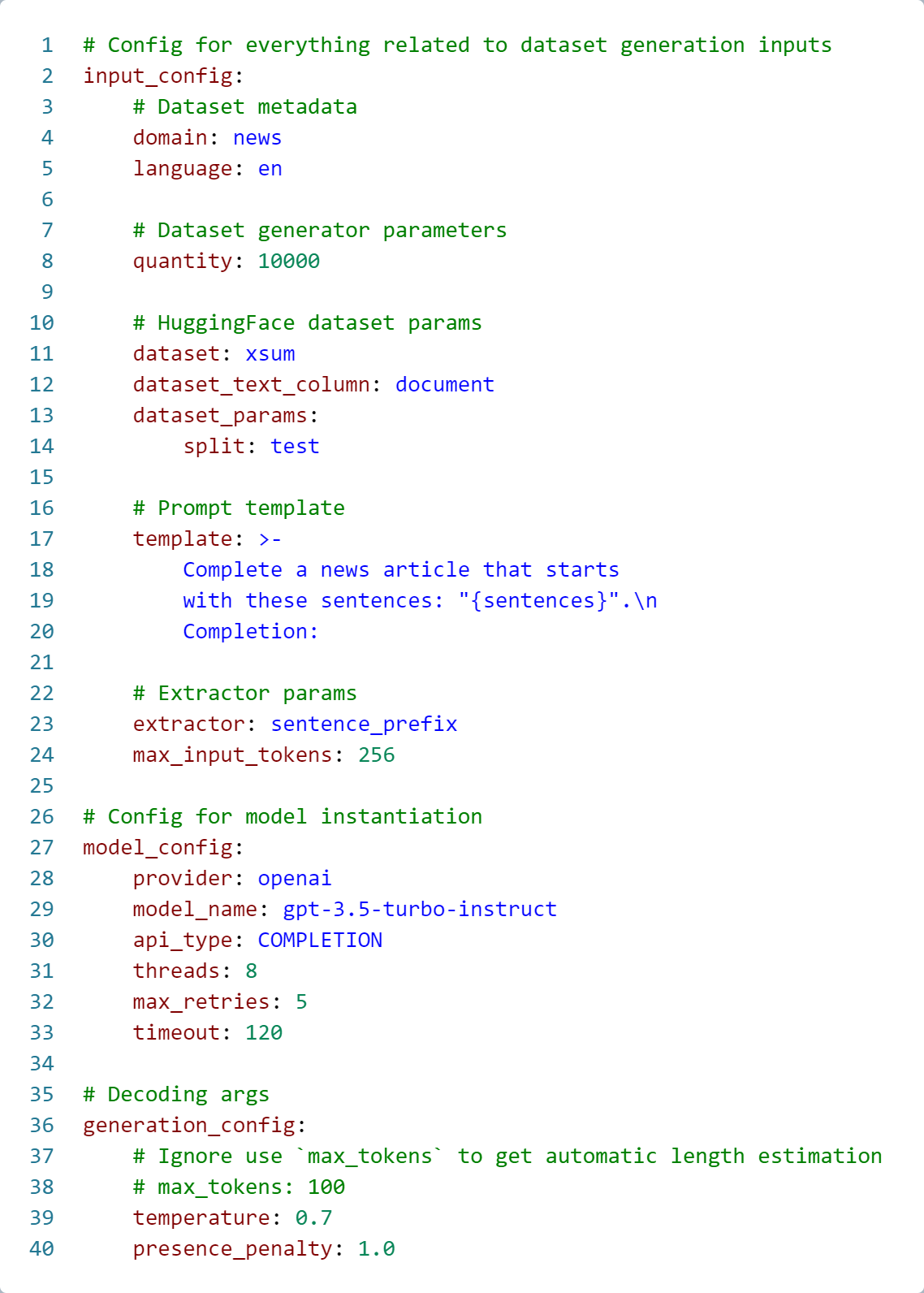}
\caption{\label{fig:boundary-config} YAML config to generate a boundary detection dataset of 10,000 news articles, using human texts from the XSum dataset, by prompting \texttt{gpt-3.5-turbo-instruct} using the \texttt{SentencePrefix} extractor.}
\end{figure*}

We show in Figure \ref{fig:boundary-config} how to define a configuration file to create a boundary detection dataset from human-written news of the XSum \cite{narayan-etal-2018-dont} dataset by prompting OpenAI's \textit{gpt-3.5-turbo-instruct}. 
Note that we use the \texttt{SentencePrefix} extractor to fill the prompt templates with the first $k$ random sentences of each article.\footnote{If $k$ is passed as argument to the extractor, it picks the first $k$ sentences. Otherwise, $k$ will be random.}
Although users can leverage any kind of extractor to build boundary detection datasets, it is highly recommended to use \texttt{SentencePrefix} or \texttt{WordPrefix} with a random $k$ to avoid biases that lead boundary detection models to count sentences or words.

\begin{figure*}[!ht]
\centering
\includegraphics[width=\linewidth]{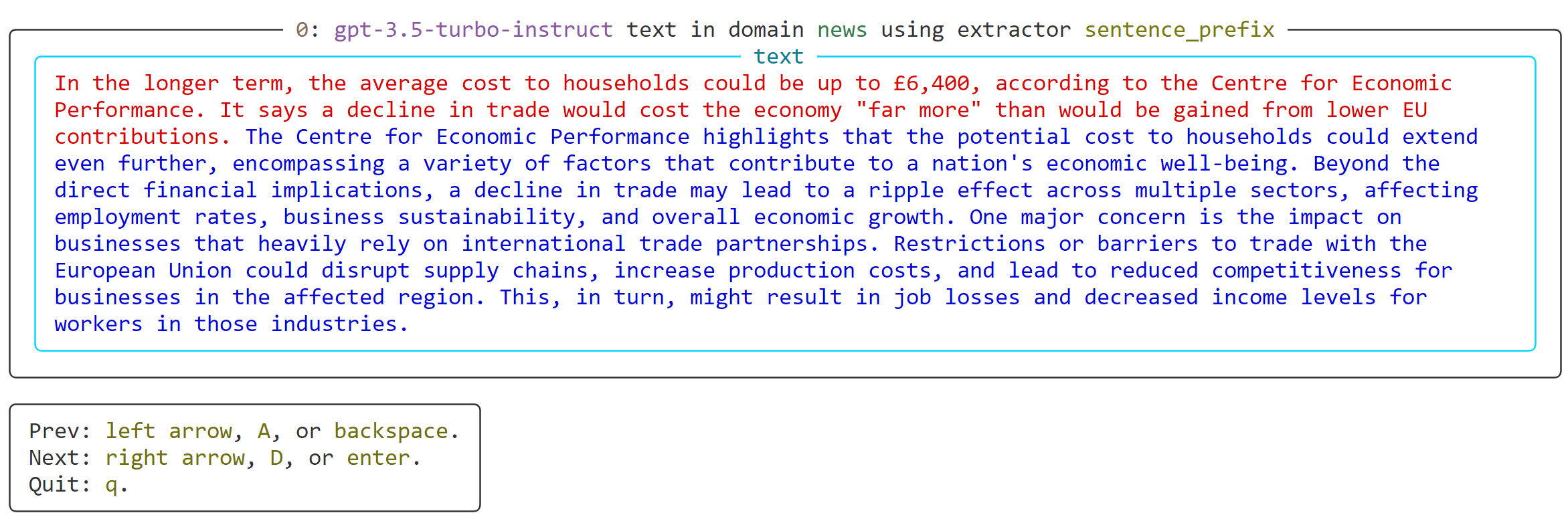}
\caption{\label{fig:cli-boundary}Interactive exploration of a boundary detection dataset.}
\end{figure*}

With the configuration file shown in Figure \ref{fig:boundary-config}, we can run the \texttt{explore} endpoint to inspect a small dataset (see Table \ref{tab:use-cases}). 
The interactive interface to inspect boundary detection datasets is shown in Figure \ref{fig:cli-boundary}. 
If the quality of the small dataset generated with that specific configuration is good enough, the user can then run the \texttt{generate} endpoint to build the boundary detection dataset.

\section{Generating a Mixcase Detection Dataset}
\begin{figure*}[!ht]
\centering
\includegraphics[scale=0.36, clip]{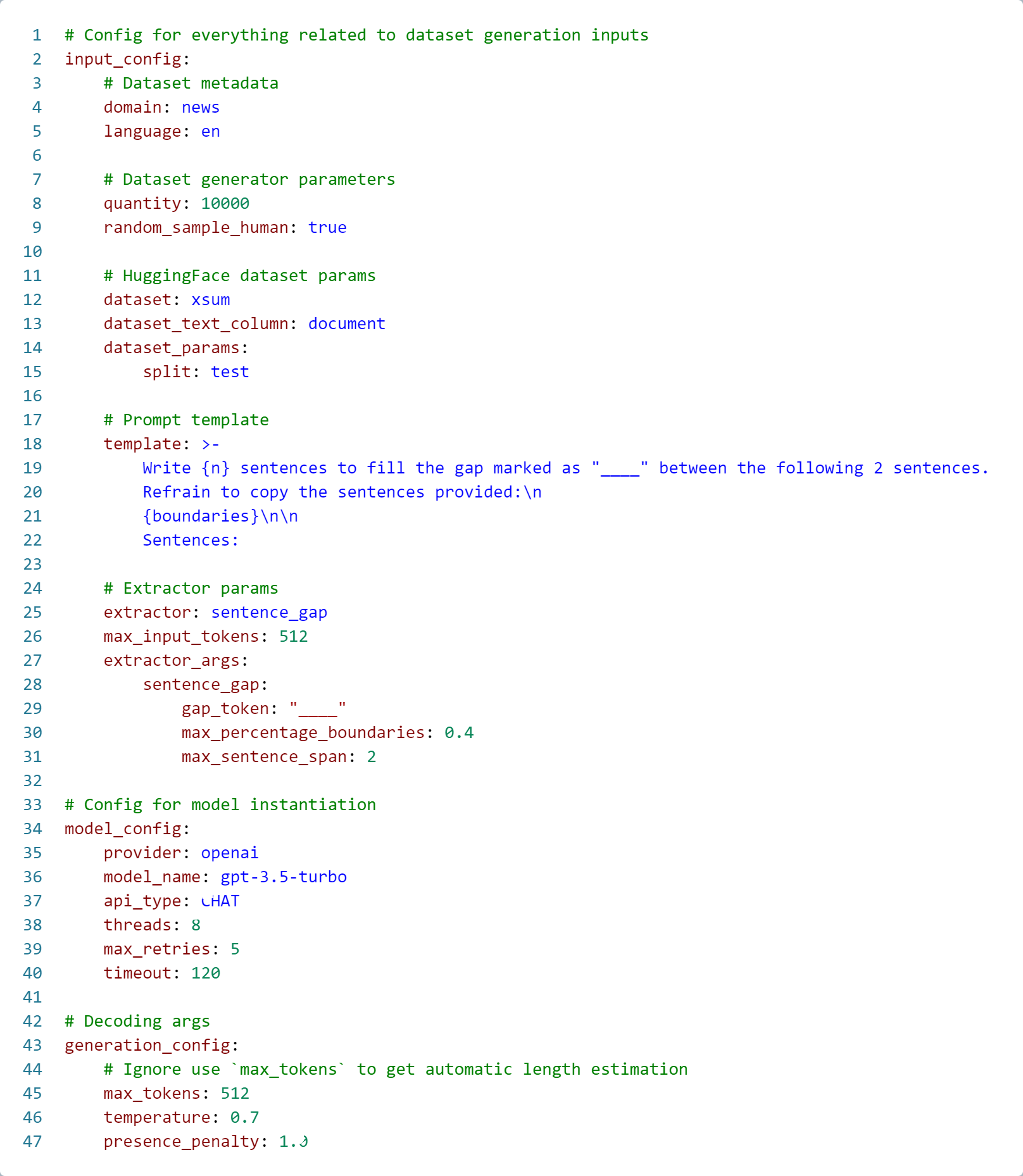}
\caption{\label{fig:mixcase-config} YAML config to generate a mixcase detection dataset of 10,000 news articles, using human texts from the XSum dataset, by prompting \texttt{gpt-3.5-turbo} using the \texttt{SentenceGap} extractor.}
\end{figure*}

\begin{figure*}[!ht]
\centering
\includegraphics[width=\linewidth]{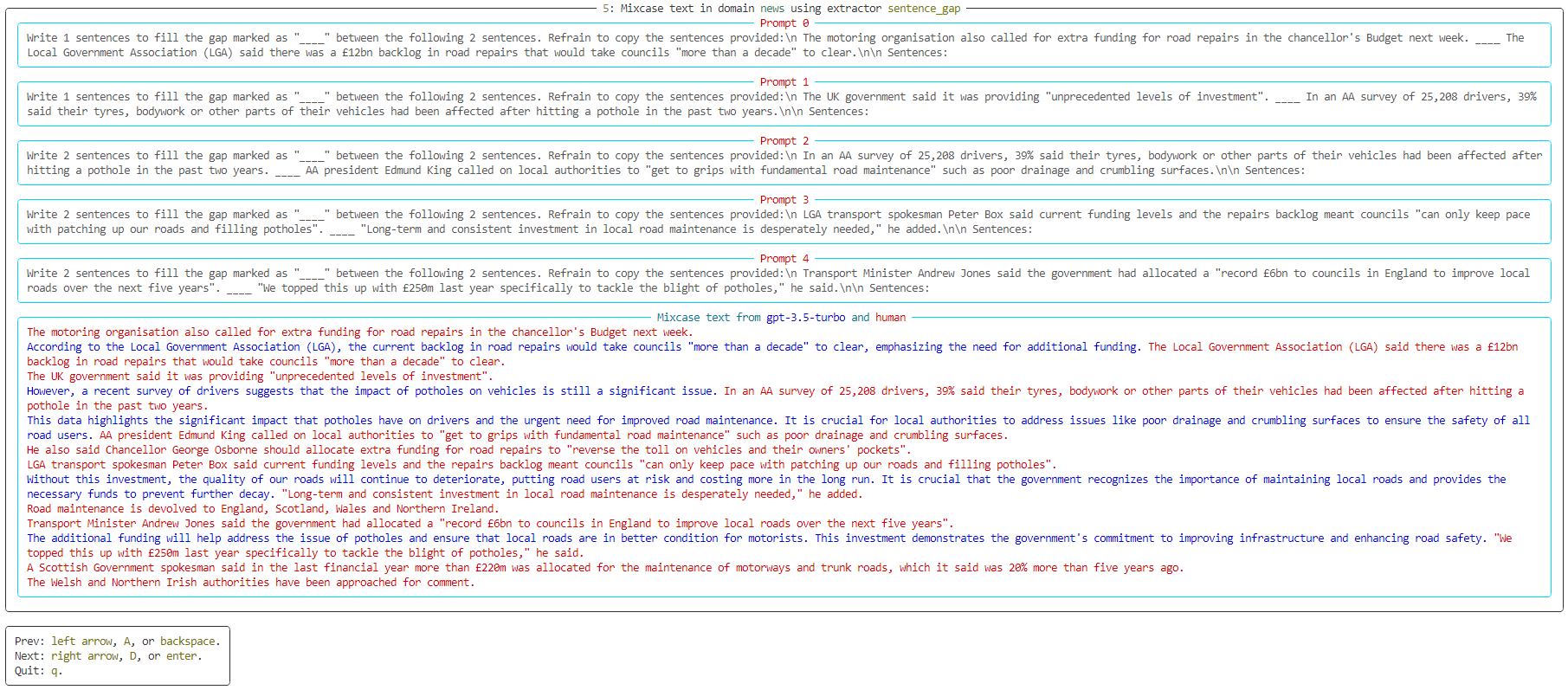}
\caption{\label{fig:cli-mixcase}Interactive exploration of a mixcase detection dataset.}
\end{figure*}

Figure \ref{fig:mixcase-config} shows a configuration file to create a mixcase detection dataset from human-written news of the XSum dataset by prompting OpenAI's \textit{gpt-3.5-turbo}. In this example, we use \textit{SentenceGap} as extractor, thus, \textmachina~randomly extracts several sentence boundaries, according to \textit{max\_percentage\_boundaries}, for \textit{gpt-3.5-turbo} to write a maximum of \textit{max\_sentence\_span} sentences within the boundaries. Finally, the generated samples consist of interleaved human and generated sentences, as shown in Figure \ref{fig:cli-mixcase}.

The previous example is one of the three possible strategies that can be used to generate mixcase detection datasets in \textmachina. The other two strategies are rewriting and masking. For rewriting, users can use the \textit{SentenceRewriting} extractor, which randomly samples several sentences of the texts to be rewritten by an LLM. For masking, the \textit{SentenceMasking} and \textit{WordMasking} extractors can be employed. These extractors randomly replace sentences or words spans by a mask token, and an LLM has to fill all the masks at once. Instead of relying only on a boundary of one preceding and one succeding sentence/word span, as in \textit{SentenceGap} and \textit{WordGap}, this masking strategy allows the LLMs to use all the text as context, thus generating more coherent text. However, this task requires to generate a JSON output and generate all the masks, which is only addressable by the highly-capable LLMs such as GPT-4.

\end{document}